\documentclass[conference]{IEEEtran}
\IEEEoverridecommandlockouts

\usepackage{url}
\usepackage{array,tabularx}
\usepackage{multicol} 
\usepackage{cite}
\usepackage{amsmath,amssymb,amsfonts}
\usepackage{graphicx}
\usepackage{algpseudocode}
\usepackage{textcomp}
\usepackage{float}
\usepackage{algorithm}
\usepackage{amsthm}
\usepackage{afterpage}
\usepackage[table]{xcolor} 
\usepackage{hyperref}
\usepackage{booktabs} 
\usepackage{caption}

\allowdisplaybreaks

\begin{document}
\title{RF-source Seeking with Obstacle Avoidance using Real-time Modified Artificial Potential Fields in Unknown Environments
\thanks{
Under review for IEEE ICCAS 2025}
}


\author{
Shahid Mohammad Mulla, Aryan Kanakapudi, Lakshmi Narasimhan, Anuj Tiwari\\
\textit{Indian Institute of Technology Madras, Chennai, India}\\
shahid1mulla2@gmail.com, aryankanna2004@gmail.com, lnt@ee.iitm.ac.in, anujt@iitm.ac.in
}

\maketitle




\begin{abstract}
Navigation of UAVs in unknown environments with obstacles is essential for applications in disaster response and infrastructure monitoring. However, existing obstacle avoidance algorithms such as Artificial Potential Field (APF) are unable to generalize across environments with different obstacle configurations. Furthermore, the precise location of the final target may not be available in applications such search and rescue, in which case approaches such as RF source seeking can be used to align towards the target location. This paper proposes a real-time trajectory planning method, which involves real time adaptation of APF through a sampling-based approach. The proposed approach utilizes only the bearing angle of the target without its precise location, and adjusts the potential field parameters according to the environment with new obstacle configurations in real time.  The main contributions of the article are i) RF source seeking algorithm to provide a bearing angle estimate using RF signal calculations based on antenna placement, and ii) modified APF for adaptable collision avoidance in changing environments, which are evaluated separately in the simulation software Gazebo, using ROS2 for communication. Simulation results show that the RF source-seeking algorithm achieves high accuracy, with an average angular error of just 1.48 degrees, and with this estimate, the proposed navigation algorithm improves the success rate of reaching the target by 46\% and reduces the trajectory length by 1.2\% compared to standard potential fields.

\end{abstract}

\section{Introduction} 
 The increasing use of drones in various applications has been facilitated by advancements in sensor technology, enabling better localization and obstacle detection methods. These technologies allow drones to effectively navigate through complex environments, avoiding obstacles in real time. The demand for autonomous drone navigation is growing in sectors like search and rescue \cite{mayer:hal-02128385}, inspection of unknown areas\cite{PANIGATI2025106101}, and other critical applications requiring drones to operate in unfamiliar and potentially hazardous environments. In these scenarios, drones must autonomously identify and locate targets, update environmental maps in real time, detect obstacles, and plan safe trajectories. The variability of these environments, such as changes in obstacle sizes, distances, and spatial constraints, poses a significant challenge to creating a unified navigation system that can adapt to such differing conditions.

The problem can be broken down into several subproblems. First, assuming localization is already handled, the direction or location of the target must be communicated to the drone. Next, real-time navigation must be executed by continuously mapping the surroundings and detecting obstacles as they appear. Finally, the varying parameters of the environment need to be accounted for, so the drone can determine an optimal trajectory in real time.

RF source seeking is an effective method for communicating the target’s location to the drone. The RF source, emitting detectable RF signals, serves as the target for the drone. For instance, in search and rescue operations, this RF source could be a cellphone of a stranded individual\cite{Tai2022-sh}, or in the case of inspection, a mobile landing site in a cluttered environment\cite{An_Kang_Choi_Lee_2023}. RF-based navigation is particularly robust in conditions where GPS is unavailable or unreliable, such as indoors, underwater, or in dense urban areas \cite{Yang2021}. 

Numerous techniques are available for calculating the direction vector or coordinates of an RF source \cite{8692423}, each with distinct trade-offs related to accuracy, hardware complexity, and computational demands:

\begin{itemize}
    \item \textbf{RSSI (Received Signal Strength Indicator):}\cite{10.1007/978-3-319-39207-3_32} This approach is straightforward to implement and cost-effective. However, it is highly susceptible to multipath fading and environmental noise, which can significantly reduce localization accuracy, particularly in non-line-of-sight conditions. RSSI provides only distance measurements, necessitating the use of multiple drones or transmitters for accurate localization.

    \item \textbf{ToF (Time of Flight) \& TDoA (Time Difference of Arrival):}\cite{von_Tschirschnitz_2019} These methods offer higher localization accuracy but require multiple receivers and transmitters, along with precise time synchronization across all devices, adding complexity to the hardware setup.

    \item \textbf{AoA (Angle of Arrival):}\cite{article} This technique yields precise directional information on the arrival angle of the RF signal. However, it requires advanced algorithms to mitigate multipath fading effects, such as the MUSIC algorithm, as well as complex hardware and directional antennas.

    \item \textbf{RSS \& AoA with Particle Filtering:}\cite{DBLP:journals/corr/abs-2012-05286} A highly accurate method involves combining RSS and AoA data through particle filtering. While this method enhances localization precision, it is computationally intensive and demands substantial processing power.
\end{itemize}

The primary algorithm selected in this study for RF source seeking is the Angle of Arrival (AoA) algorithm, chosen for its simplicity in requiring only a single RF source or beacon along with receivers in the form of antennas. AoA can provide a direction vector sufficient for guiding navigation using Artificial Potential Fields (APFs). Typically, research on AoA focuses on reducing multipath fading effects and enhancing the precision of the arrival angle, with approaches such as the MUSIC algorithm\cite{1143830}, which requires an antenna array to improve accuracy. Other research on AoA centers on antenna configurations for 3D source localization, focusing on implementing AoA estimation techniques in three-dimensional environments \cite{5286580}. 

However, this paper specifically emphasizes antenna configurations that extend AoA sensing capabilities. While conventional AoA setups typically employ an antenna array with a limited 180-degree sensing range in a 2D plane, the approach here aims to achieve a full 360-degree sensing range by using different antenna configurations. This configuration enables accurate and comprehensive coverage, allowing for efficient RF source seeking with reduced computational demands.

Artificial Potential Fields method (APFs) is commonly employed for navigation \cite{100007}, but it is faced with problems related to Goal Non-Reaching due to Obstacle Nearness (GNRON)\cite{doi:10.1177/0142331218824393} and local minima traps\cite{131810}, where the drone can become stuck between obstacles before reaching the target. Several studies have tackled these local minima problems using advanced variations of APF \cite{Park2003},\cite{1225434}. One approach for applying APF to real-time navigation is by dynamically updating the repulsive potential as new obstacles are detected, and updating the attractive potential using a temporary target always at a distance from the drone towards the real target. 

Many papers discuss methods to optimize the trajectory after it is generated by an APF, like \cite{4651091},\cite{9189655}. Even though the final trajectory is optimized based on an objective function, the parameters of the potential function itself need to be tuned for one particular map based on experimental results.

Some studies adapt the potential function in real-time based on environmental changes. For example, one paper uses evolutionary genetic algorithms to modify the potential function dynamically \cite{870304}. However, it is required to process fitness functions for many generations till a suboptimal solution is reached every time step. Other studies on usage of artificial intelligence to optimize potential fields include \cite{Elkilany2020}  which uses Fuzzy inference systems to optimize the parameters of potential field function in the context of formation control of ground robots and \cite{Furferi2016}  uses Artificial Neural Networks(ANNs) to obtain a relation between trajectory and potential function parameters to optimize the parameters to output desired trajectory.

A sampling-based control method called Model Predictive Path Integral (MPPI)\cite{7487277} gained popularity recently in the field of path planning of autonomous vehicles due to its flexibility in defining the costs or risks. MPPI typically introduces disturbances to control inputs to generate many trajectories, each having its own cost, and a weighted average of control inputs and respective trajectory costs is used to generate the optimal trajectory. \cite{https://doi.org/10.48550/arxiv.2308.00914} uses MPPI to perform path planning in a UAV with a decreased sampling space achieved by trajectory parametrization. A similar method is proposed in this paper to optimize the potential function parameters. This new approach introduces disturbances to the parameters of the potential field function and then calculates potential trajectories based on these modifications, with each trajectory evaluated based on a cost function considering path length, collision avoidance, smoothness, and error in reaching the final target. The parameters of the trajectory closest to the optimal trajectory are set as new function parameters.

This solution provides a complete package to autonomously guide the drone to an RF source target. It enables the drone to navigate through diverse map types without needing any modifications or human intervention along the way.

Therefore, the contributions of this work include a robust algorithm for calculating the direction of RF source with full 360 degrees sensing range using the antenna configuration; an implementation of a modified Artificial Potential Field (APF) with real-time mapping, where the temporary target is updated via the AoA-derived direction vector and a sampling-based approach to dynamically optimize potential field parameters in real-time environments.

\section{RF Source Seeking}

This section presents the complete formulation and implementation of an algorithm that employs the Angle of Arrival (AoA) method to estimate the real-time direction of an RF signal-emitting source. The algorithm is designed to capture the phase differences in received signals between the dipole antennas to compute the arrival angle, which will then be used to obtain the direction vector. This direction vector is then integrated into the modified Artificial Potential Fields (APFs) framework for obstacle avoidance and path planning which will be discussed in the next sections. 

Hence, the overall section is divided into an introduction for the Angle of Arrival, followed by a discussion on the number of antennas used for accurate signal acquisition. This section then details the process of estimating the direction vector based on the multiple AoA measurements obtained from multiple array of antennas. Finally, the implementation of the algorithm is explained using a square antenna configuration for a simple implementation with reasonable accuracy. Later, there will be separate sections which will discuss about setup of the simulation which was used  to test this algorithms.

\subsection{Angle of Arrival (AoA))}

The Angle of Arrival (AoA) estimation technique determines the direction from which an RF signal arrives at a sensor array by analyzing the phase difference across multiple antennas \cite{article}. This approach exploits the principle that a signal arriving at an angle reaches each antenna with a slight delay, resulting in measurable phase shifts.

\begin{figure}[H]
    \centering
    \includegraphics[width=0.7\linewidth]{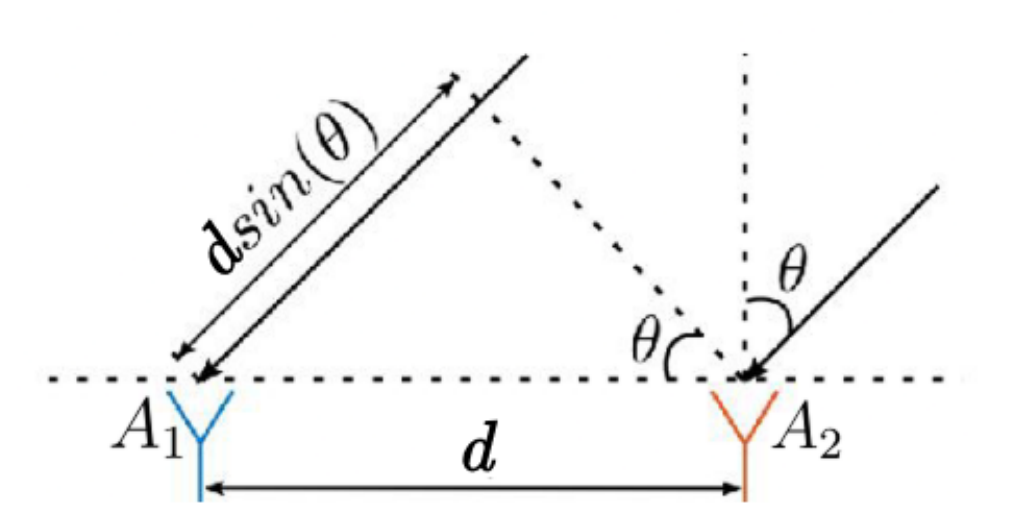}
    \caption{Angle of Arrival in case of 2 antennas.}
    \label{fig:1}
\end{figure}

A common implementation uses a Uniform Linear Array (ULA), where the $n$th antenna receives the incoming signal:
\begin{align}
s_n(t) = A \cdot e^{j(\omega t + n\phi)} + w(t)  
\end{align}
Here, \(A\) is amplitude, \(\omega\) is the carrier frequency, \(\phi\) is the AoA-induced phase shift, and \(w(t)\) the noise.
To extract the phase, the received signals are correlated with a local carrier frequency of \(\omega\), digitized and processed to get $\phi$.
Techniques like MUSIC (Multiple Signal Classification) are used to obtain this phase. MUSIC works by separating useful signal information from noise, leading to more accurate angle estimates \cite{1143830}.
For two antennas spaced by distance \(d\), the AoA \(\theta\) can be estimated as:
\begin{align}
\theta = \sin^{-1}\left(\frac{\phi \cdot \lambda}{2\pi d}\right)
\end{align}
where \(\phi\) is the phase difference and \(\lambda\) is the wavelength of the RF signal.


\subsection{Antenna Configuration for Full Range Sensing}

\begin{figure}[htbp]
    \centering
    \includegraphics[width=0.7\linewidth]{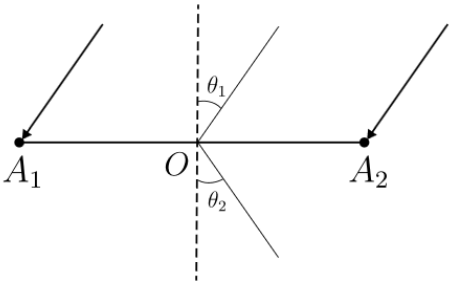}
    \caption{Two Angles of Arrival in case of 2 antennas.}
    \label{fig:3}
\end{figure}

The previously discussed algorithms exhibit an inherent limitation of a \(180^\circ\) detection range within a two-dimensional plane. This constraint arises due to the computation of the angle of arrival (AoA) using the \( \sin^{-1} \) function, which produces values in the range \( -\pi/2 \) to \( +\pi/2 \). Consequently, the detection range is restricted to a \(180^\circ\) span, effectively limiting detection to one particular side of the antenna array, as illustrated in Figure \ref{fig:3}.

If an AoA, \(\theta\), is detected, it remains ambiguous whether the source resides on the positive or negative side of the \(y\)-axis. In other words, the ambiguity arises between two potential AoA values, \(\theta_1\) and \(\theta_2\), which, while mathematically equivalent, correspond to distinct source positions.

So, to achieve full \(360^\circ\) coverage, the configuration must incorporate multiple antenna dipoles, with the \(360^\circ\) field either divided equally or unequally among these dipoles, based on the requirements of the algorithm.
An important consideration is the number of antennas required to achieve this configuration. While three non collinear antennas could theoretically suffice, a more practical approach involves using four antennas placed at specific positions, where each antenna belongs to only one linear array. This configuration simplifies the analysis and avoids shared antennas between the two linear arrays. The proposed algorithm in this paper could also be adapted to use three antennas by overlapping two antennas to form a single coinciding pair which is discussed in the appendix.

In the following derivation, it is assumed that a four-antenna configuration is utilized. After obtaining the AoA from each linear array, the next step is to identify the precise half-plane from which the RF signal originates. This half-plane determination, which will be addressed in a subsequent section after the derivation of the direction vector, is crucial for constructing an accurate directional estimate.
Once the origin side is identified, the algorithm will compute the direction vector \& coordinates of the RF source relative to the drone’s frame of reference using the AoA measurements obtained from each independent linear antenna array.

\begin{figure}{htbp}
    \centering
    \includegraphics[width=1.0\linewidth]{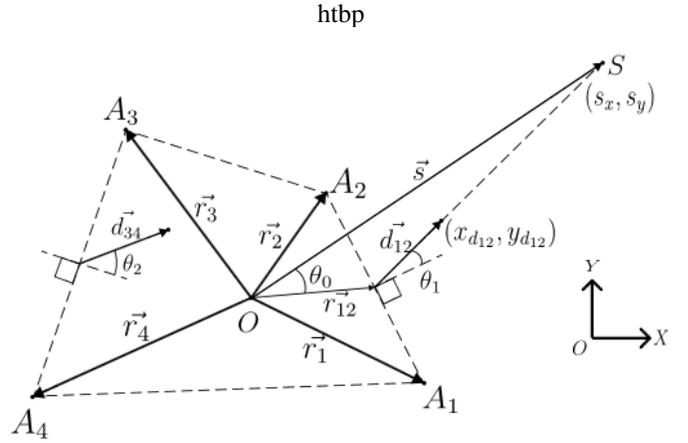}
    \caption{Arbitrary Antenna Configuration with 4 Antennas.}
    \label{fig:4}
\end{figure}

To establish a consistent notation and clarify the system's configuration, consider Figure \ref{fig:4}. Let \( A_i \) represent the \(i^{th}\) antenna, positioned at \(\vec{r}_i\) relative to the drone’s center of mass for all \( i \in \{1, 2, 3, 4\} \). The RF source, denoted by \( S \), has a position vector \(\vec{s}\). The angles of arrival (AoA) are given by \(\theta_1\) for dipole \( A_1A_2 \) and \(\theta_2\) for dipole \( A_3A_4 \).

Now, there are a few constraints that should be taken into consideration before starting the derivation:

1. \textbf{Antenna Spacing}:\cite{Doan2023-yt} To accurately capture phase differences in received signals, the spacing between antennas within each dipole should not exceed half the wavelength of the incoming signal, i.e., 
\begin{align}
   |\vec{r}_i - \vec{r}_j| \leq \frac{\lambda}{2}
\end{align}

   where \((i, j) = (1, 2)\) \& \((3, 4)\) correspond to the antenna pairs in each dipole.

2. \textbf{Dynamic Stability}: To maintain stability, it is essential that the placement of antennas does not introduce unbalanced forces, assuming equal antenna weights. This balance is achieved by ensuring that the vector sum of all antenna positions equals zero:
\begin{align}
     \vec{r}_1 + \vec{r}_2 + \vec{r}_3 + \vec{r}_4 = \vec{0}
\end{align}

3. \textbf{Non-collinearity Constraint}: To avoid reducing the system to a single linear antenna array, the four antennas must not be collinear. Mathematically, this condition is imposed by ensuring that:
   \begin{align}
       \frac{(\vec{r}_i - \vec{r}_j) \cdot (\vec{r}_k - \vec{r}_l)}{\|\vec{r}_i - \vec{r}_j\| \|\vec{r}_k - \vec{r}_l\|} \neq \pm 1
   \end{align}
\noindent  
where \((i, j, k, l)\) are chosen such that each pair represents distinct antenna vectors.

To proceed with the derivation of the antenna placement algorithm, several variables will be defined to facilitate the calculation of the RF source’s position. Let \(\vec{r}_{34}\) represent the midpoint of antennas \(A_3\) and \(A_4\), given by:
\begin{align}
    \vec{r}_{34} = \frac{\vec{r}_3 + \vec{r}_4}{2}
\label{abcd}
\end{align}
Similarly, let \(\vec{r}_{12}\) represent the midpoint of antennas \(A_1\) and \(A_2\), calculated as:
\begin{align}
    \vec{r}_{12} = \frac{\vec{r}_1 + \vec{r}_2}{2}
\end{align}
These midpoints serve as reference points for the angle of arrival (AoA) measurements at each linear array.
 \(\vec{d}_{34}\) is defined as the unit direction vector obtained from the AoA at \(\vec{r}_{34}\) and \(\vec{d}_{12}\) as the unit direction vector from the AoA at \(\vec{r}_{12}\). Now, the position of the RF source \(\vec{s}\) can be represented as:
\begin{align}
    \vec{r}_{34} + k_{34} \vec{d}_{34} = \vec{s}
\end{align}
\begin{align}
    \vec{r}_{12} + k_{12} \vec{d}_{12} = \vec{s}
\end{align}
\noindent
where \(k_{34}\) and \(k_{12}\) are scaling factors applied to the unit vectors \(\vec{d}_{34}\) and \(\vec{d}_{12}\) respectively, to complete the vector geometry from each midpoint to the source. The values of \(k_{34}\) and \(k_{12}\) are unknown and will be determined subsequently.

By substituting \(\vec{r}_{34}\) and \(\vec{r}_{12}\) with their definitions, the equations for \(\vec{s}\) become:
\begin{align}
    \frac{\vec{r}_3 + \vec{r}_4}{2} + k_{34} \vec{d}_{34} = \vec{s}
\end{align}
\begin{align}
    \frac{\vec{r}_1 + \vec{r}_2}{2} + k_{12} \vec{d}_{12} = \vec{s}
\end{align}
These equations can be expanded in terms of the \(x\)- and \(y\)-coordinates as follows:
\begin{align}
    \frac{x_3 + x_4}{2} + k_{34} \left( x_{d_{34}} - \frac{x_3 + x_4}{2} \right) = s_x
\end{align}
\begin{align}
    \frac{y_3 + y_4}{2} + k_{34} \left( y_{d_{34}} - \frac{y_3 + y_4}{2} \right) = s_y
\end{align}
\begin{align}
    \frac{x_1 + x_2}{2} + k_{12} \left( x_{d_{12}} - \frac{x_1 + x_2}{2} \right) = s_x
\end{align}
\begin{align}
    \frac{y_1 + y_2}{2} + k_{12} \left( y_{d_{12}} - \frac{y_1 + y_2}{2} \right) = s_y
\end{align}
\noindent
where:\\
- \((x_i, y_i)\) denotes the coordinates of \(\vec{r}_i\) \(\forall i \in \{1, 2, 3, 4\}\),\\
- \((s_x, s_y)\) denotes the coordinates of \(\vec{s}\),\\
- \((x_{d_{34}}, y_{d_{34}})\) denotes the endpoint coordinates of \(\vec{d}_{34}\),\\
- \((x_{d_{12}}, y_{d_{12}})\) denotes the endpoint coordinates of \(\vec{d}_{12}\).\\
These four equations can be written in the matrix form as a system of linear equations:
\begin{align}
\begin{bmatrix}
1 & 0 & 0 & \frac{x_3 + x_4}{2} - x_{d_{34}} \\
0 & 1 & 0 & \frac{y_3 + y_4}{2} - y_{d_{34}} \\
1 & 0 & \frac{x_1 + x_2}{2} - x_{d_{12}} & 0 \\
0 & 1 & \frac{y_1 + y_2}{2} - y_{d_{12}} & 0 \\
\end{bmatrix}
\begin{bmatrix}
s_x \\
s_y \\
k_{12} \\
k_{34}
\end{bmatrix}
=
\begin{bmatrix}
\frac{x_3 + x_4}{2} \\
\frac{y_3 + y_4}{2} \\
\frac{x_1 + x_2}{2} \\
\frac{y_1 + y_2}{2}
\end{bmatrix}
\end{align}
This system is of the form \( Ax = b \). The solution to this system is given by \(x = A^{-1} b\). For a unique solution to exist, the determinant of \( A \) must be non-zero, i.e., \(\det(A) \neq 0\).

Once a solution for the system is obtained, it provides both the scaling constants \( k_{12} \) \& \( k_{34} \) and the coordinates of the RF source, \((s_x, s_y)\). Direction vector of the source relative to the origin can then be  calculated using these coordinates:
\begin{align}
    \vec{d}_{s} = \frac{\vec{s}}{\|\vec{s}\|}
\end{align}

But, if the \( \det(A) \) becomes zero, the system \( x = A^{-1}b \) no longer yields a unique solution. This situation can result in two distinct cases: either there is no solution, or there are infinitely many solutions which are discussed in detail along with figures in the Appendix.

\begin{figure}[htbp]
    \centering
    \includegraphics[width=0.7\linewidth]{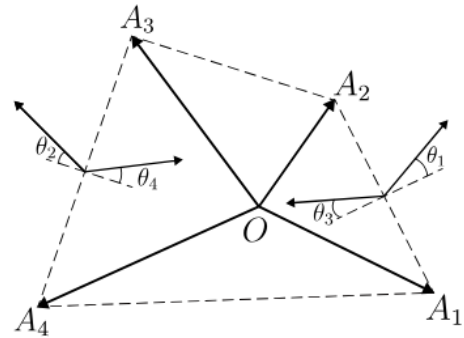}
    \caption{Two Angles of Arrival for the dipole \(A_1A_2\) \& \(A_3A_4\).}
    \label{fig:9}
\end{figure}

Now, consider the dipole \(A_1A_2\) in Figure \ref{fig:9}. As discussed above, a key challenge here is distinguishing whether the received RF signals are originating from the direction \( \theta_1 \) or \( \theta_3 \). The derivation above operates under the assumption that the signals approach from direction \( \theta_1 \), allowing for the computation of the vector \( \vec{d}_{12} \). However, distinguishing mathematically between \( \theta_1 \) and \( \theta_3 \) is not straightforward in implementation, especially in configurations where antennas are arranged arbitrarily. So, creating a conditional algorithm to determine the source direction for any general arrangement can be complex and computationally demanding.

One effective approach to resolve the ambiguity in source direction is through cross-correlation\cite{Callaghan2011-fb}, a method well-suited for identifying the signal’s originating side by measuring the relative phase or time delay of received signals between paired antennas. Cross-correlation allows for the determination of which antenna experiences a time lag, which in turn reveals the general direction of arrival. For example, applying cross-correlation between pairs such as \( A_2 \) and \( A_3 \) or \( A_1 \) and \( A_4 \) will indicate the source’s relative position by showing which antenna in the pair receives the signal first.

In this paper, while the concept of cross-correlation is acknowledged, the primary focus will be on developing a mathematical framework to determine the correct angle of arrival (AoA) for RF signals specifically within a square antenna configuration. The square configuration is chosen for its computational simplicity, symmetry, and compatibility with the constraints previously outlined.

\subsection{Square Antenna Configuration}

The proposed methodology is now applied to a square antenna configuration. The approach involves calculating the inverse of the matrix \(A\) and deriving the final direction vector in terms of the angles \(\theta_1\) and \(\theta_2\). To validate the correctness of the derived equations, the solution obtained using this method is compared with the solution derived using basic geometric principles. The comparison serves to verify the accuracy of the mathematical formulation for a simplified antenna arrangement. 

\begin{figure}[htbp]
    \centering
    \includegraphics[width=0.8\linewidth]{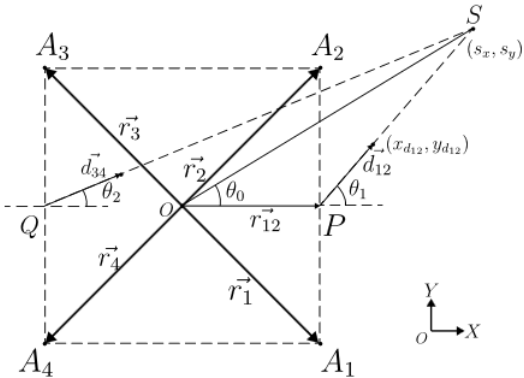}
    \caption{Square Antenna Configuration.}
    \label{fig:10}
\end{figure}

The square configuration is specifically chosen as it represents the primary proposal for the antenna arrangement due to its symmetry, computational simplicity \& simplification of implementation of the algorithm. Referring to Figure \ref{fig:10}, the configuration is depicted as a square arrangement of antennas, with notations consistent with those in Figure \ref{fig:4} and the distance between 2 adjacent antennas being \(2d\). This arrangement also satisfies the constraints outlined in previous sections, thereby making it a practical choice for real-world applications.

Before proceeding with the estimation of the final direction vector, it is necessary to first determine the half-plane from which the RF signals are arriving. This can be achieved by mathematically comparing the two angles of arrival, \(\theta_1\) and \(\theta_2\), which are obtained at the dipoles \(A_1A_2\) and \(A_3A_4\), respectively. 
\begin{itemize}
    \item If \(\theta_1 > \theta_2\), it implies that the source is located on the half-plane of the dipole pair \(A_1A_2\) (right side of O).

    \item Conversely, if \(\theta_2 > \theta_1\), the source is located on the half-plane of \(A_3A_4\) (left side of O).
\end{itemize}

While the difference between these angles may be small due to the assumption that the source is located at a significant distance for AoA calculation, this angular discrepancy can still be used to determine the appropriate half-plane for the source.

Once the half-plane is determined, the next step is to proceed with the estimation of the final direction vector. To simplify the equations, we substitute the coordinates of the square configuration as follows: 
\begin{align}
    (x_1, y_1) = (d, -d), \quad (x_2, y_2) = (d, d)
\end{align}
\begin{align}
 (x_3, y_3) = (-d, d), \quad (x_4, y_4) = (-d, -d),
\end{align}
Which leads to the following simplifications:
\begin{align}
    \left(\frac{x_1 + x_2}{2}\right) = d, \quad \left(\frac{x_3 + x_4}{2}\right) = -d, 
\end{align}
\begin{align}
    \left(\frac{y_1 + y_2}{2}\right) = 0, \quad \left(\frac{y_3 + y_4}{2}\right) = 0.
\end{align}
Furthermore, the expressions for the direction vectors \( \vec{d}_{12} \) and \( \vec{d}_{34} \) are given by:
\begin{align}
    x_{d_{12}} = d + \cos\theta_1, \quad y_{d_{12}} = \sin\theta_1, 
\end{align}
\begin{align}
    x_{d_{34}} = -d + \cos\theta_2, \quad y_{d_{34}} = \sin\theta_2.
\end{align}
Substituting these values into the equation \(A x = b \), derived in the previous section, results in a simplified \( A \) matrix and \( b \) vector:
\begin{align}
A = \begin{bmatrix}
1 & 0 & 0 & -\cos\theta_2 \\
0 & 1 & 0 & -\sin\theta_2 \\
1 & 0 & -\cos\theta_1 & 0 \\
0 & 1 & -\sin\theta_1 & 0
\end{bmatrix}, \quad
b = \begin{bmatrix}
-d \\
0 \\
d \\
0
\end{bmatrix}.
\end{align}

The determinant of matrix \( A \) is computed as:
\begin{align}
    \det(A) = \sin(\theta_1 - \theta_2),
\end{align}
Which implies that if \(\det(A) = \sin(\theta_1 - \theta_2) = 0\), then \(\theta_1 = \theta_2\), leading to parallel direction vectors or the case where the midpoints of the dipole pairs \(A_1A_2\) and \(A_3A_4\) become collinear with the source, as discussed previously. In such cases, the system has no unique solution or infinitely many solutions, respectively. 

If \(\det(A) \neq 0\), the solution for \( x \) is obtained by computing the inverse of matrix \( A \):
\begin{align}
    x = A^{-1} \cdot b
\end{align}
\begin{align}
\begin{bmatrix} s_x \\ s_y \\ k_{12} \\ k_{34} \end{bmatrix}
= \begin{bmatrix}
-d \, cosec(\theta_2 - \theta_1) (\sin\theta_2 \cos\theta_1 + \cos\theta_2 \sin\theta_1) \\
-2d \, cosec(\theta_2 - \theta_1) \sin\theta_1 \sin\theta_2 \\
-2d \, cosec(\theta_2 - \theta_1) \sin\theta_2 \\
-2d \, cosec(\theta_2 - \theta_1) \sin\theta_1
\end{bmatrix}.
\end{align}
From the above solution, the coordinates \((s_x, s_y)\) of the RF source can be determined, and the direction vector \( \vec{d}_s \) from the origin is given by:
\begin{align}
    \tan\theta_0 = \frac{s_y}{s_x}
\end{align}
Upon simplification, this results in the following expression for \(\theta_0\):
\begin{align}
    \tan\theta_0 = \frac{2 \sin\theta_1 \sin\theta_2}{\sin\theta_2 \cos\theta_1 + \cos\theta_2 \sin\theta_1}
\end{align}
Finally, inverting the tangent expression yields:
\begin{align}
    \cot\theta_0 = \frac{\cot\theta_1 + \cot\theta_2}{2}
\end{align}
This expression provides the estimated angle \(\theta_0\), which represents the angle of arrival of the RF signals with respect to the origin of the coordinate system.

Now, to verify the formulation presented above, the sine rule is applied to triangles \( QSO \) and \( OSP \) in Figure \ref{fig:10}. By equating the side \( OS \) in both sine rule equations, the following relationship is obtained:
\begin{align}
    \frac{\sin\theta_2}{\sin(\theta_0 - \theta_2)} = \frac{\sin\theta_1}{\sin(\theta_1 - \theta_0)}
\end{align}
Upon expanding and rearranging the terms, the following equation is derived:
\begin{align}
    \cot\theta_0 = \frac{\cot\theta_1 + \cot\theta_2}{2}
\end{align}
This result confirms that the algorithm provides the correct direction vector \( \vec{d}_s \) for the source, regardless of its location. 

The rationale for adopting this algorithm, even though the calculations for the square configuration were relatively simple, lies in its flexibility. This method can be generalized to accommodate arbitrary antenna configurations and can be extended to any number of antennas. Moreover, even estimation techniques can be employed to further reduce the error by incorporating additional antenna array, thereby improving the accuracy of the direction vector estimation.

\subsection{Final Algorithm in a Square Antenna Configuration}

The following algorithm outlines the final methodology for RF source localization using a square antenna configuration. The mathematical steps derived earlier are incorporated into the pseudo-code for clarity and implementation.

\begin{algorithm}[H]
\caption{\textsc{RF Source Seeking}}
\textbf{Input:} 
\(\theta_1, \theta_2\) (Angles of arrival), \(d\) (Half-length of the square's side).

\textbf{Initialize:}
\begin{itemize}
    \item Midpoints: \( M_{12} = (d, 0), \; M_{34} = (-d, 0) \).
    \item Direction vectors: 
    \[
    (x_{d_{12}}, y_{d_{12}}) = (d + \cos\theta_1, \sin\theta_1),
    \]
    \[
    (x_{d_{34}}, y_{d_{34}}) = (-d + \cos\theta_2, \sin\theta_2).
    \]
\end{itemize}

\textbf{Run:}
\begin{algorithmic}[1]
\State Identify the source half-plane:
\If{$\theta_1 > \theta_2$}
    \State Source lies in the half-plane of \( A_1A_2 \).
\Else
    \State Source lies in the half-plane of \( A_3A_4 \).
\EndIf

\State Compute:
\[
\cot\theta_0 = \frac{\cot\theta_1 + \cot\theta_2}{2}.
\]

\State Output the final direction vector:
\[
\vec{d}_s = \begin{bmatrix} \cos\theta_0 \\ \sin\theta_0 \end{bmatrix}.
\]
\end{algorithmic}
\end{algorithm}

\section{Navigation in Unknown Environments}

After receiving the direction of the RF signal as a vector, the drone has to detect locally observable obstacles in real time and navigate through these obstacles in the RF source seeking direction simultaneously. Real-time local mapping can be achieved by any existing Simultaneous Location and Mapping (SLAM) techniques and is not a focus of this paper. The drone should then use this partial local map of the environment to plan a path to a horizon until the map updates with newer obstacles being detected. The method of artificial potential fields (APFs) is one of the standard approaches to plan a safe path avoiding obstacles due to its high safety and simplicity. This method can be adapted to real-time applications by updating these potential functions based on the map being updated at every time step. However, this method will not work for every map due to the local minima trap \cite{131810} \& GNRON problem \cite{doi:10.1177/0142331218824393}, and the parameters will need to be tuned for every environment the drone encounters to avoid these problems. So, to avoid this parameter tuning a sampling based method inspired by Model Predictive Path Integral (MPPI) Control is adopted such that these parameters can be tuned real time and the potential well will be updated based on the requirements like varying density of obstacles, etc which will be discussed in detail in the later sections.

Therefore, the following sections briefly discuss the main idea of artificial potential fields and a method to utilize artificial potential fields in real time, problems faced by this method in detail, which is then followed by a basic introduction to MPPI Control and the discussion on optimization of APFs using a method inspired by MPPI Control. 

\subsection{Artificial Potential Fields}
In the artificial potential field method, the robot will experience an attractive force from a target point and repulsive forces from all the obstacles, and the resultant of these forces guides the robot towards the goal while also avoiding obstacles. These forces are modeled as electrostatic forces that appear if the robot and obstacles have the same charge and the goal has the opposite charge. This arrangement can also be viewed as if there is a potential field formed due to these charges of goal \& obstacles, and the robot is always trying to move in a path with the least potential to reach the goal. The potential due to a goal is called attractive potential, and the potential due to obstacles is called repulsive potential, and when combined, forms the complete potential field on which gradient descent is applied to calculate this path with the least potential.

Equations for attractive potential and repulsive potential are as follows:
\\The attractive potential function is given by:

\begin{align}
    U_{\text{att}}(\mathbf{q}) = \frac{1}{2} k_{\text{att}} \|\mathbf{q} - \mathbf{q}_{\text{goal}}\|^2
\end{align}
\noindent
where:
\begin{itemize}
    \item $\mathbf{q}$ is the current position of the drone.
    \item $\mathbf{q}_{\text{goal}}$ is the goal position.
    \item $k_{\text{att}}$ is the attraction constant.
    \item $\|\cdot\|$ denotes the Euclidean norm.
\end{itemize}

The repulsive potential function is defined as:

\begin{align}
    U_{\text{rep}}(\mathbf{q}) = \sum_{i=1}^{n} k_{\text{rep}} \left( \frac{1}{d_{\text{boundary}}(i)}-\frac{1}{d_0} \right)
\end{align}

Where:
\begin{itemize}
    \item $n$ is the number of obstacles.
    \item $\mathbf{q}$ is the current position of the drone.
    \item $\mathbf{q}_{\text{obstacle}}^{(i)} = (x_{\text{obstacle}}, y_{\text{obstacle}}, r_{\text{obstacle}})$ is the $i$-th obstacle's position and radius.
    \item $d_{\text{boundary}}(i)$ is the distance from the drone to the boundary of the $i$-th obstacle, given by:
    \begin{align}
        d_{\text{boundary}}(i) = \|\mathbf{q} - \mathbf{q}_{\text{obstacle}}^{(i)}\| - r_{\text{obstacle}} - r_{\text{drone}}
    \end{align}
    where $r_{\text{drone}}$ is the radius of the drone.
    \item $d_0$ is the influence range within which the repulsive potential is effective.
    \item $k_{\text{rep}}$ is the repulsion constant.
\end{itemize}

In this paper, a logarithm-based repulsive potential is implemented instead of the traditional function.
The new repulsive potential function is given by:
\begin{align}
    U_{\text{rep}}(\mathbf{q}) = \sum_{i=1}^{n} \left(-k_{\text{rep}} \log\left( \frac{d_{\text{boundary}}(i)}{d_0} \right)\right)
\end{align}

The gradient of the logarithm-based repulsive potential still shows variations at relatively large values of distances, unlike the usual function, ensuring higher safety and decreasing the likelihood of the gradient becoming zero.\cite{WANG2024104052}

The total potential function can be calculated as:

\begin{align}
U_{\text{total}}(\mathbf{q}) = U_{\text{att}}(\mathbf{q})+U_{\text{rep}}(\mathbf{q})
\label{Eq_U_total}
\end{align}

\begin{figure}[htbp]
    \centering
    \includegraphics[width=0.7\linewidth]{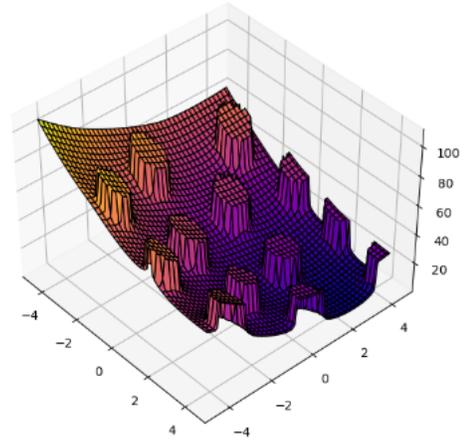}
    \caption{An example potential function.}
    \label{fig:11}
\end{figure}

The total potential at a point is least when the point is the goal and increases the farther the point is from the goal, with obstacles being high potential points as shown in figure \ref{fig:11}. The drone should move towards the direction along which the potential drops the most to reach the goal while avoiding obstacles. Hence, the desired trajectory is calculated by applying a gradient descent algorithm with the drone position as the starting point, which can be tracked by any trajectory control.

Till now, it is assumed that the location of all obstacles and the final goal are known from the start. If the environment is unknown and the drone is mapping the obstacles as it goes, the potential needs to be updated as well. When new obstacles are revealed, they can be fed into the repulsive potential function and hence update the total function iteratively. And in this particular problem, if only the direction of the source is known, a temporary target is introduced to calculate the attractive potential. The temporary target is placed at a fixed distance along the direction of the source, making the global minima of the potential as this temporary target and trajectory is generated using the gradient descent algorithm. This temporary target can then be updated whenever the drone reaches the nth point of the trajectory, and this generates a new trajectory from this point. This process is done repeatedly until the drone reaches its final target.

\subsection{Optimization of APFs}

It is already been discussed that using artificial potential fields may lead to problems like local minima traps and GNRON, but parameters, mainly the attractive coefficient, the repulsive coefficient, and the influence range, can be tuned to avoid such problems. This tuning needs to be done manually for every unique map, running the process multiple times. This can be challenging in an unknown environment, as not every part of the environment is mapped in each run while tuning the parameters. The environment might have different levels of obstacle density at different places, and parameters that are best for one initial position might not be useful at all for a different initial position. And parameters with which potential field works fine till half of a run might suddenly start causing problems in the remaining part of the map if the map starts to get more cluttered. Even if the current parameters make the drone reach the desired target point, the path the drone took might not be a desired path in terms of length, safety, energy usage, or any other factors that are required.

These problems call for a method to tune these parameters autonomously and in real time, such that the drone takes a desired path every time the temporary goal gets updated. As there is no direct way to determine the correct parameters in real time that would produce a trajectory with desired performance criteria, search-based optimization techniques can be utilized with a constructed cost function to find parameters that would produce an optimal or at least a suboptimal trajectory. This paper proposes a method similar to Model Predictive Path Integral (MPPI) control, a sampling-based motion planning algorithm. MPPI usually needs less number of samples and only one iteration to find a solution, making it perfect for solving a real-time optimization problem. 

The overall methodology is divided into 3 phases:
\begin{enumerate}
    \item Parameter Sampling
    \item Trajectory evaluation
    \item Optimal trajectory
\end{enumerate}

These 3 phases of the algorithm run at every temporary target update, generating a new trajectory to track each time.
\subsubsection{Parameter Sampling}

The parameters that affect the potential field and ultimately the trajectories generated are: attractive potential coefficient, repulsive potential coefficient, and the distance from which the potential affects the drone. Every time the temporary target gets updated, an initial trajectory is generated with the initial values of this set of parameters, and then the parameters are repeatedly sampled for a specific number of times, and a corresponding potential field is generated for each set. These potential fields each give a trajectory reaching the temporary target, which are evaluated using a cost function. 

The three parameters are randomly chosen from a Gaussian Distribution with a fixed mean and standard deviation for each parameter. 
\begin{align}
    k_{\text{att}_{\text{new}}} \sim \mathcal{N}(\mu_{k_{\text{att}}}, \sigma^2_{\text{att}})
\end{align}
\begin{align}
    k_{\text{rep}_{\text{new}}} \sim \mathcal{N}(\mu_{k_{\text{rep}}}, \sigma^2_{\text{rep}})
\end{align}
\begin{align}
    {d_0}_{\text{new}} \sim \mathcal{N}(\mu_{d_0}, \sigma^2_{d_0})
\end{align}
These parameters are sampled until all the values are strictly above their corresponding set limits. The sampling count till which these parameters are sampled depends on the computational power available. 

\subsubsection{Trajectory Evaluation}
Once many sets of parameters are sampled, their potential fields are generated, and trajectories are calculated. Each trajectory is evaluated using a cost function depending on the requirements. For now, the cost function consists of penalties on the length of the trajectory, error of the final point of the trajectory in reaching the temporary target, the number of sharp turns, and proximity to obstacles.
Assuming that the trajectory consists of \(n\) points \((x_i, y_i)\) for \(i = 1, 2, \dots, n\). 
The cost function for the trajectory can be expressed as:

\begin{align}
    \text{Cost} = L + E + A + P
\end{align}

Where:

\textbf{1. Length of Trajectory (\(L\)):}

The total length of the trajectory is given by the sum of the distances between consecutive points:

\begin{align}
    L = \sum_{i=1}^{n-1} \sqrt{(x_{i+1} - x_i)^2 + (y_{i+1} - y_i)^2}
\end{align}

\textbf{2. Error to Temporary Goal (\(E\)):}

The error to the temporary target is the Euclidean distance between the final point of the trajectory \((x_n, y_n)\) and the temporary target \((x_t, y_t)\):

\begin{align}
    E = \sqrt{(x_n - x_t)^2 + (y_n - y_t)^2}
\end{align}

\textbf{3. Angle Deviations (\(A\)):}

The angle deviations are the sum of the absolute differences in angles between consecutive trajectory points:

\begin{align}
    A = \sum_{i=2}^{n-1} \lvert \theta_i - \theta_{i-1} \rvert
\end{align}

Where \( \theta_i \) is the angle between consecutive points of the trajectory.

\textbf{4. Proximity to Obstacles (\(P\)):}

The proximity penalty is given by:

\begin{align}
    P = \frac{1}{\min_{i,j} \text{dist}((x_i, y_i), \text{Obstacle}_j)}
\end{align}

Where \( \text{dist}((x_i, y_i), \text{Obstacle}_j) \) is the distance between the trajectory point \((x_i, y_i)\) and the \(j\)-th obstacle.

\subsubsection{Optimal Trajectory}
After calculating all the trajectories along with cost functions, an optimal trajectory is generated by calculating a weighted average of these trajectory points. The weights of each trajectory are calculated based on the cost function of each trajectory as follows:
\begin{align}
    w = \exp\left(-\frac{S}{\lambda}\right)
\end{align}

Where \(\lambda\) is a constant. This particular formulation helps assign weights so that trajectories with higher cost will have lower weights and trajectories with lower cost will have higher weights. This ensures that trajectories with lower cost value get higher priority. This method is chosen over selecting the trajectory with the least cost because it provides flexibility in calculating the optimal trajectory. The constant \(\lambda\) can be increased or decreased depending on the required strictness of the weights. \(\lambda\) can either be increased to make the weights stricter to ignore the trajectories with slightly higher cost value or decreased to make the weights more relaxed, making the trajectories with slightly higher cost values contribute to the optimal trajectory significantly.
The optimal trajectory is computed as the initial trajectory plus the weighted average of distances from the initial trajectory to all sampled trajectories.

\begin{align}
    T_{\text{opt}}(i) = T_{\text{init}}(i) + 
\frac{\sum_{j=1}^n w_j \cdot \lvert T_j(i) - T_{\text{init}}(i) \rvert}{\sum_{j=1}^n w_j}
\end{align}

Where:
\begin{itemize}
    \item \(T_{\text{opt}}(i)\): The \(i\)-th point of the optimal trajectory.
    \item \(T_{\text{init}}(i)\): The \(i\)-th point of the initial trajectory.
    \item \(T_j(i)\): The \(i\)-th point of the \(j\)-th sampled trajectory.
    \item \(w_j\): The weight of the \(j\)-th sampled trajectory.
     \item \(\lvert T_j(i) - T_{\text{init}}(i) \rvert\): The Euclidean distance between the points on the initial trajectory and the \(j\)-th sampled trajectory.
\end{itemize}

\begin{algorithm}
\caption{\textsc{trajectory optimization Algorithm}}
\textbf{Input:} 
\begin{itemize}
    \item Initial position: \(\mathbf{P}_0\), Goal position: \(\mathbf{P}_g\), Obstacles: \(\mathcal{O}\).
    \item Parameters: \(k_{\text{att}}\) (attraction), \(k_{\text{rep}}\) (repulsion), \(d_0\) (obstacle influence distance).
    \item Maximum steps: \(N_{\text{steps}}\), Sampling count: \(N_{\text{sample}}\), Weight decay factor: \(\lambda\).
\end{itemize}

\textbf{Run:}
\begin{algorithmic}[1]
\State Initialize the temporary target as \(\mathbf{P}_{\text{temp}} = \mathbf{P}_g\).

\While{Drone has not reached \(\mathbf{P}_g\)}
    \If{Drone reaches waypoint near \(\mathbf{P}_{\text{temp}}\)}
        \State Update \(\mathbf{P}_{\text{temp}}\) based on intermediate progress.
    \EndIf
    
    \State Compute initial trajectory \(\mathcal{T}_{\text{initial}}\) from \(\mathbf{P}_0\) to \(\mathbf{P}_{\text{temp}}\) using \(N_{\text{steps}}\).
    
    \State Initialize empty lists for sampled trajectories, objective values, and weights.
    
    \For{\(i = 1\) to \(N_{\text{sample}}\)}
        \State Sample parameters \((k_{\text{att}}', k_{\text{rep}}', d_0')\):
        \[
        k_{\text{att}}' \sim \mathcal{N}(k_{\text{att}}, \sigma_{\text{att}}), \;
        k_{\text{rep}}' \sim \mathcal{N}(k_{\text{rep}}, \sigma_{\text{rep}}) \; (\text{ensure } k_{\text{rep}}' > limit_{kr}), 
        \]
        \[
        d_0' \sim \mathcal{N}(d_0, \sigma_{d_0}) \; (\text{ensure } d_0' > limit_{d_0}).
        \]
        
        \State Compute trajectory \(\mathcal{T}_i\) using sampled parameters.        
        \State Compute Cost \(J_i\) for Trajectory \(\mathcal{T}_i\)
        \State Compute weight \(w_i = e^{-J_i / \lambda}\).
        
        \State Append \(\mathcal{T}_i\), and \(w_i\) to respective lists.
    \EndFor
    
    \State Compute the optimal trajectory \(\mathcal{T}_{\text{optimal}}\):
    \[
    \mathcal{T}_{\text{optimal}}[t] = \mathcal{T}_{\text{initial}}[t] + \frac{\sum_i w_i \cdot \|\mathcal{T}_i[t] - \mathcal{T}_{\text{initial}}[t]\|}{\sum_i w_i}.
    \]
    
    \State Identify the closest trajectory \(\mathcal{T}_{\text{closest}}\) to \(\mathcal{T}_{\text{optimal}}\):
    \[
    \mathcal{T}_{\text{closest}} = \arg\min_i \sum_t \|\mathcal{T}_i[t] - \mathcal{T}_{\text{optimal}}[t]\|.
    \]
    
    \State Feed \(\mathcal{T}_{\text{closest}}\) to the trajectory tracking function.
\EndWhile
\end{algorithmic}
\end{algorithm}

As this is an average of trajectories, there is no guarantee that this trajectory does not go through obstacles, even though there is a penalty related to proximity to obstacles in the cost function. But there is guarantee that none of the sampled trajectories go through obstacles, as these are generated using the gradient descent algorithm on a potential field. Hence, a sampled trajectory closest to this optimal trajectory in terms of distance is chosen as the final trajectory to be followed by the drone. 

This whole process with 3 stages is repeated when the drone reaches an ith point in the set of way points of the trajectory to generate a new trajectory to reach the new temporary goal. When the new temporary goal is within a range of the final goal, the temporary goal is set as the final goal, and the last trajectory is generated, completing the journey.

\section{Simulations}
Multiple simulations were conducted to test and validate the previously discussed algorithms in various types of environments. The drone simulation is taken from the SJTU drone simulation program, which utilizes Gazebo with ROS2 for control. The drone, modeled after the AR Drone, is equipped with cameras, sonar, GPS, and odometry can be directly accessed. In these experiments, the drone is controlled by giving velocity commands in Cartesian directions relative to the drone, and these velocity commands are controlled using a PID controller, and this same controller is used in every simulation, testing out and comparing the algorithms in various environments.

To simulate real-time mapping, a full point cloud representing the entire designed environment was generated as the ground truth using gazebo\_map\_creator. The drone was programmed to only perceive the points visible within a virtual cone representing its field of view, constructed based on the drone's current position and orientation. Points falling within this cone were extracted, transformed using coordinate transformations, and mapped onto a global frame. As the drone navigated the environment, it incrementally revealed the map, simulating the process of real-time mapping. The obstacles for the algorithm are extracted by considering the points of the point cloud at the same height as the drone, and a circle fitting algorithm is used to calculate the position of the center and approximate radius of each obstacle with respect to world coordinates. All the point cloud manipulation in this simulation is done using the Open3D library.

Simulation of RF signals is achieved by using a standard \(2^{nd}\) order wave equation that varies over the x, y, and z coordinates and over time. This wave equation is given by:

\begin{align}
    W(x, y, z, t) = A_i \sin \left[ k .d_i + \omega (t - t_0) + \phi \right]
\end{align}
Where \(d_i\) is given by:
\begin{align}
    d_i = \sqrt{(x_i - x_0)^2 + (y_i - y_0)^2 + (z_i - z_0)^2}
\end{align}
The coordinate \((x_0,y_0,z_0)\) is the position of the RF source, which is different in each scenario of every environment. This wave equation is computed at the four antenna positions fixed on top of the drone in a square configuration. The computed value is then fed to the Fast Fourier Transform to calculate the phase values, which will then be used for the AoA algorithm. If the above-discussed algorithm is being implemented on a real drone, it is recommended to pass the signal through the MUSIC algorithm or any other likewise algorithm to avoid problems related to multi-path propagation. After the direction vector of the RF source is calculated, this vector is then fed to the Navigation stack to define a temporary target.

During the experiments, the initial position of the drone, chosen arbitrarily, is recorded as the starting position of the trajectory. The total region is defined as a fixed 10×10 m square grid for all simulations, ensuring that the initial position, target, and obstacles reside within this area in all cases. The total potential is calculated at every point in the grid with a resolution of 0.5 m. Each scenario features a distinct arrangement of obstacles with varying levels of density and complexity. The proposed navigation algorithm is tested on these scenarios along with standard Artificial potential fields for comparison.

The position data needed for the algorithm is subscribed from the odometry topic of the drone simulation. With drone position and mapped obstacles, the temporary target is located at a distance of 2 meters, and the trajectories are generated with 15 waypoints each. This is achieved by setting a limit on the maximum number of steps for gradient descent can be calculated. The temporary target is updated when the drone reaches the 5th waypoint. The sampling count of parameters is decided to be 10, generating 10 different trajectories for the optimization. Once the final trajectory is decided, a trajectory tracking control is implemented to chase the next waypoint iteratively. The error between the drone's current position and the trajectory's next waypoint in global planar coordinates is fed into PID controllers tuned according to the drone's behavior to get global velocities. These velocities are then converted into drone coordinates using the yaw of the drone and published to the velocity command topic. 

The potential field is plotted in a heat map and a 3D plot, and the position of the drone, final target, temporary target, sampled trajectories, optimal trajectory, final trajectory, and the total path covered by the drone are plotted in the heat map.

In all simulations, the drone completes its mission by triggering an end sequence. At the final trajectory point, the drone detects a specific landing pad which is a red circle placed at the target using its downward-facing camera. 

A dedicated stopping algorithm is implemented to manage this process. Once the red circle is detected in the camera frame, the navigation algorithm terminates. Using OpenCV, the algorithm identifies the red circle and estimates the position of the landing pad's center relative to the drone's camera center. A PID controller minimizes this distance by aligning the drone with the center of the landing pad. Once the drone is stabilized, the code for plotting all the results is triggered.

\section{Results}

This section presents the results obtained from evaluating the algorithms proposed in the above section. The evaluation is mainly based on how robust the algorithm is under different environments in generating a successful trajectory and how much better it performs as compared to a standard APF. So, there will be two subsections which will evaluate the accuracy of the source seeking method and how well the modified APF performs with the errors passed on in the direction vector from the source seeking method.

\subsection{RF Source Estimation}
This subsection presents the results related to RF source estimation. The algorithm was tested in two distinct simulated scenarios to evaluate its accuracy.

\begin{figure}[htbp]
    \centering
    \includegraphics[width=0.45\textwidth]{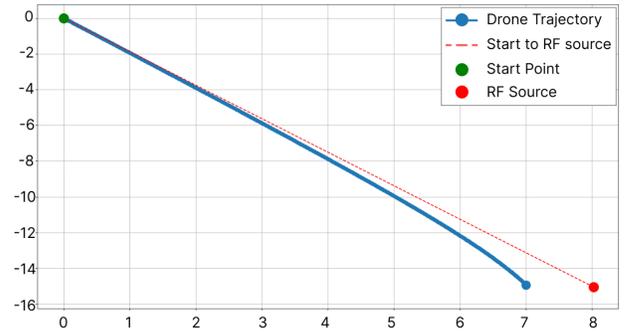}
    \caption{Drone trajectory during RF source seeking.}
    \label{fig:trajectory}
\end{figure}

In the first scenario, the drone was tasked with locating the RF source starting from a random initial position. Then, the computed direction vector was used as a control input to steer the drone in its direction. The trajectory of the drone was tracked until it terminated at the landing sequence. The trajectory visualization is shown in Figure~\ref{fig:trajectory}. And based on the trajectory of the drone, the average error calculated till the drone initiated the landing sequence was \(1.48^o\).

\begin{figure}[htbp]
    \centering
    \includegraphics[width=0.45\textwidth]{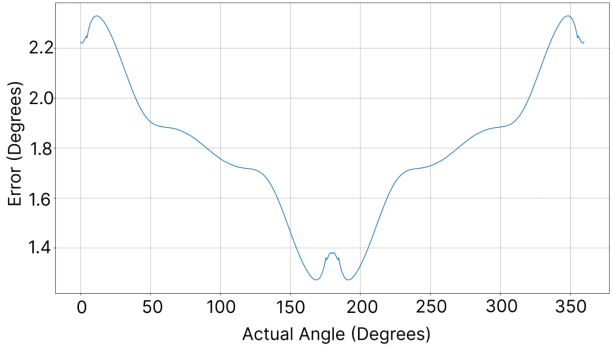}
    \caption{Error in angle estimation at various positions around the drone.}
    \label{fig:error_plot}
\end{figure}

In the second test, the RF source was moved in a circular path around the drone at a fixed radius of 5 meters. The drone was kept stationary at the center, and the algorithm's angle estimation was evaluated at various positions around the circle. The error was calculated as the difference between the angle estimated by the algorithm and the actual angle computed using the drone's and source's coordinates. The error plot is presented in Figure~\ref{fig:error_plot}. While this error is very low as compared to other standard techniques, it has to be taken into account that this algorithm has only been tested in a simulation; there will be larger errors when testing in a real experiment.

\subsection{Trajectory Planning}

\begin{figure}[htbp]
    \centering
    \includegraphics[width=0.48\textwidth]{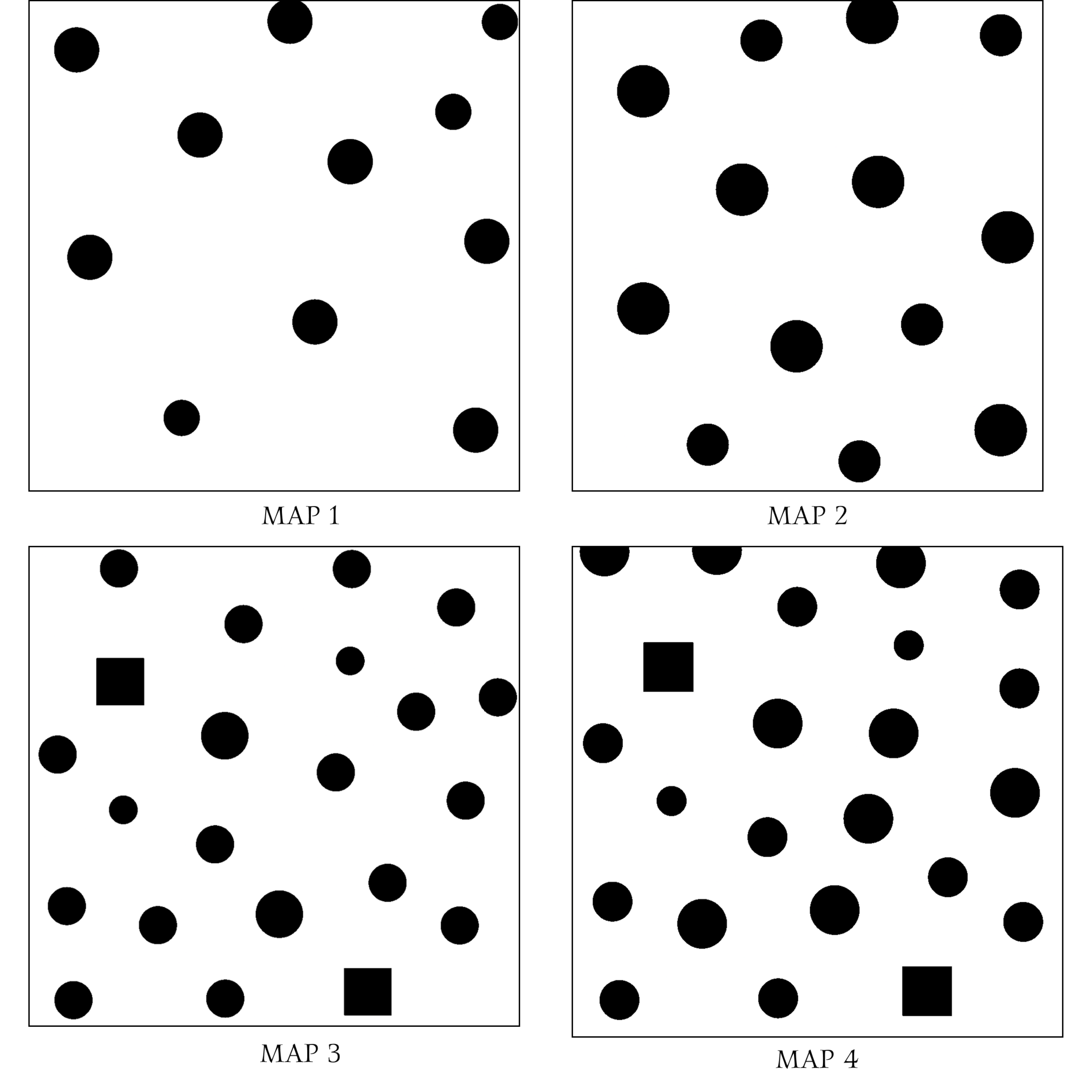}
    \caption{Four maps with uniform obstacle density.}
    \label{fig:maps}
\end{figure}

\begin{figure}[htbp]
    \centering
    \includegraphics[width=0.45\textwidth]{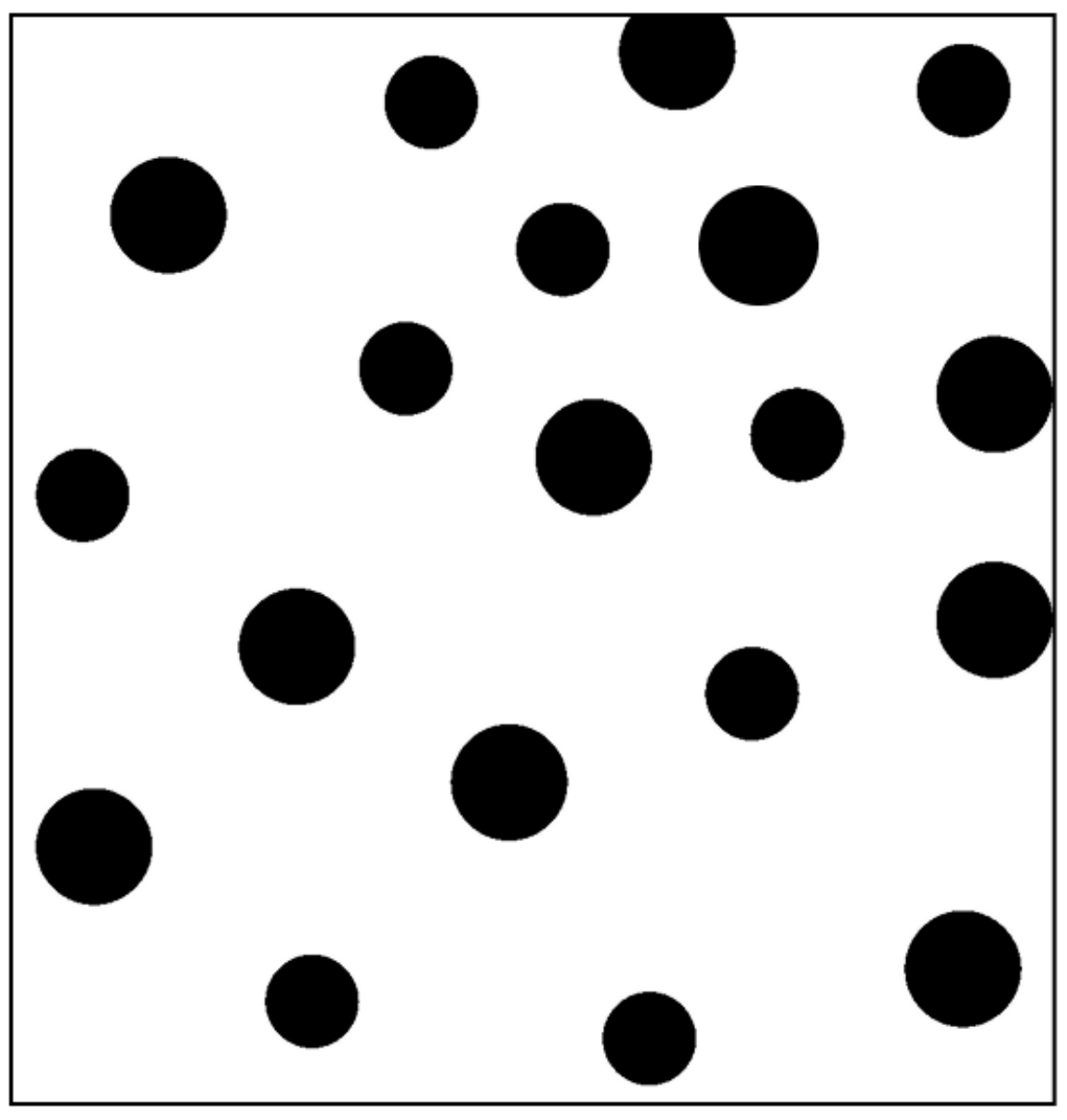}
    \caption{Map with varying obstacle density.}
    \label{fig:var}
\end{figure}

To truly compare the adaptability of the new method, the two algorithms are used to generate trajectories in five different environments. These five environments are designed based on a metric called obstacle density. It is defined as the area of all obstacles per unit square divided by the total area per unit square. The maps are compared based on the average obstacle densities in all the unit squares. Four of the five maps have constant obstacle density throughout the map as shown in figure ~\ref{fig:maps}. This can be verified by the low variance of obstacle densities in these maps compared to the last map, which is designed to have a higher variance by making the obstacle density drastically change in space as shown in figure ~\ref{fig:var} to evaluate the real-time adaptability of the algorithm. The mean and variances of five maps are shown in the table~\ref{tab:map_density}.

\begin{table}[h!]
\centering
\rowcolors{2}{gray!10}{white}
\begin{tabular}{|l|c|c|}
\hline
\rowcolor{gray!30}
\textbf{Map} & \textbf{Mean Obstacle Density} & \textbf{Variance} \\
\hline
Map 1 & 0.0702 & 0.0020 \\
Map 2 & 0.1102 & 0.0023 \\
Map 3 & 0.1267 & 0.0021 \\
Map 4 & 0.1520 & 0.0011 \\
Map 5 & 0.1552 & 0.0152 \\
\hline
\end{tabular}
\caption{Mean obstacle density and variance for all maps}
\label{tab:map_density}
\end{table}

Both algorithms generate trajectories on all five maps seven times each with randomly chosen initial and target positions. The initial parameters for all these runs are set to be the parameters obtained by experimentally tuning the standard APF to work on map 2.

Success rate and average relative length are chosen to compare these simulations. Success rate is the percentage of successful runs out of seven runs. Average relative length is defined as the average of all seven lengths, with each length divided by the straight line distance between the corresponding initial and target positions. The relative length of a particular run is taken into the final average only when both algorithms generate a valid trajectory using those initial and target positions. The final plots of success rates and average relative lengths of the four maps are plotted as shown in the figures.

\begin{figure}[htbp]
    \centering
    \includegraphics[width=0.48\textwidth]{success_rate_plot.pdf}
    \caption{Success rates of standard APF and modified APF in various environments. A video comparison of the standard APF and the proposed modified APF in Map~3 is available \href{https://drive.google.com/file/d/1T066YyVPM6804Kp4Bqs4_t3MB3fWjsvo/view?usp=sharing}{\color{blue}{here}}.}
    \label{fig:success_rate}
\end{figure}

\subsubsection{Maps with constant obstacle density}
It can be observed in the figure ~\ref{fig:success_rate} that as the obstacle density increases, the success rate for both algorithms decreases as it becomes more and more difficult to navigate through obstacles. But the success rate of modified APF remains higher for all four maps. It can be observed that the success rate plot of standard APF has a peak at map 2 because the parameters are tuned for standard APF to work in map 2, as mentioned earlier.

\begin{figure}[htbp]
    \centering
    \includegraphics[width=0.48\textwidth]{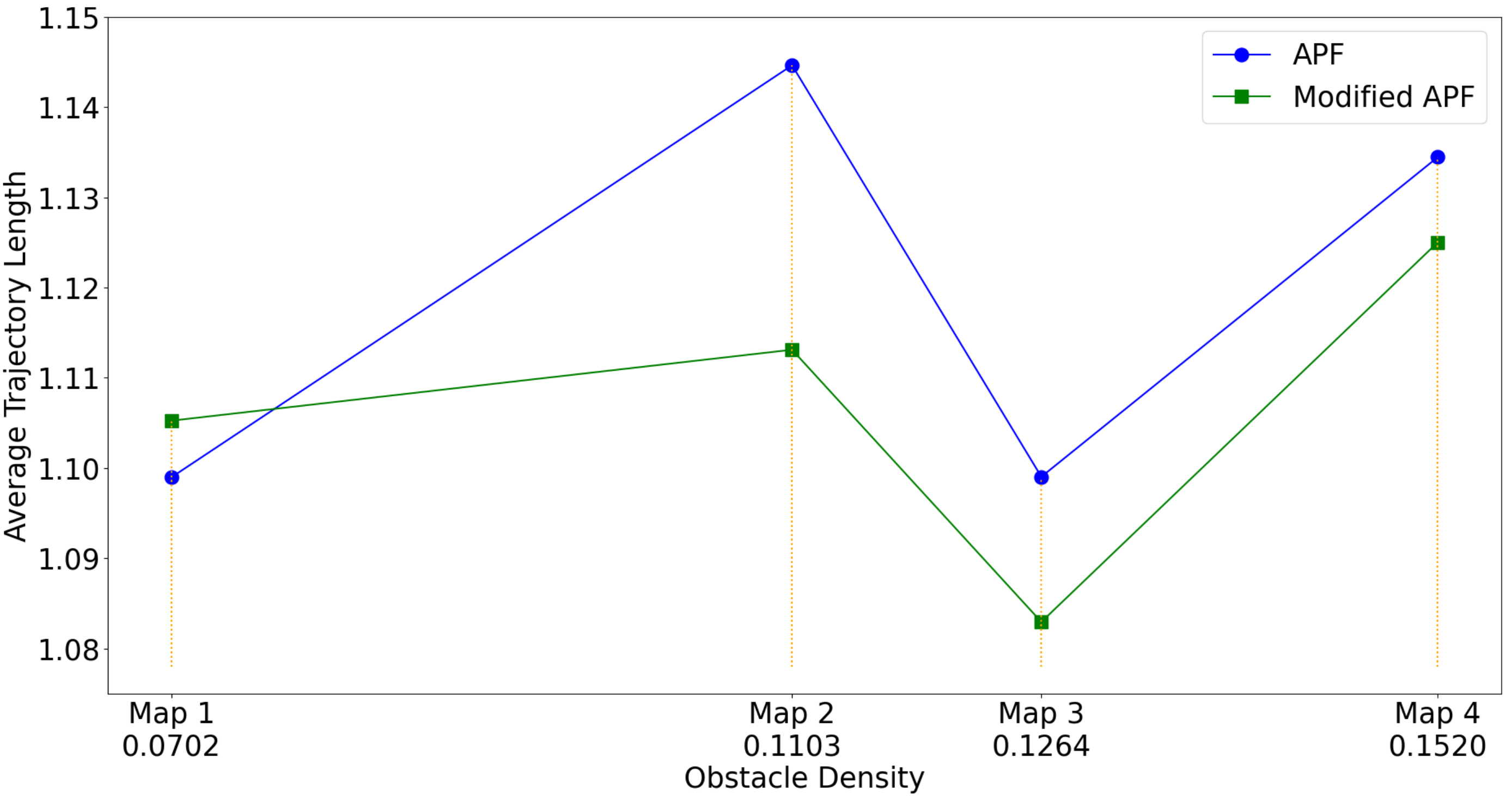}
    \caption{Average relative lengths of standard APF and modified APF in various environments.}
    \label{fig: lent}
\end{figure}

From the figure~\ref{fig: lent} it can be seen that modified APF performs better in almost all maps, generating shorter paths than standard APF on average, except in the map with the least obstacle density. This small deviation can be due the the nature of sampling. When sample trajectories are generated in a map, there is a possibility that all samples are worse,i.e., longer than the trajectory generated using standard APF at the same point. This possibility is higher in case of the map with very low obstacle density, as there is more space for samples to explore without being influenced by obstacles, compared to a map with higher obstacle density in which the sample trajectories are already constricted into limited spaces because of many obstacles, hence generating similar and in most cases better samples. This can lead to the final average length for modified APF being slightly higher than standard APF for very low density map as we can observe.

\subsubsection{Maps with varying obstacle density}

To evaluate real-time adaptability of Modified APF, the initial and target points were chosen in a well-distributed manner such that a few trajectories would go from higher obstacle density to lower and vice verse for other few, rest were made to navigate from low to low and high to high obstacle densities. Standard APF couldn't generate a trajectory in all seven runs in the map with varying obstacle density due to fixed parameter values. Modified APF managed to achieve a success rate of 57.143\% proving that it can adapt to the real-time changes in obstacle density. 

\section{Conclusion}

This research presents a new path planning algorithm based on APFs, primarily focusing on its adaptability and robustness to static changes in the environment like obstacle density while also optimizing its path length and only using the direction vector as an input to navigate towards the goal. The Modified APF algorithm introduces random sampling to overcome local minima, showing smoother and safer paths in simulation. Though tested only with circular obstacles, it can be extended to handle complex shapes. Its performance depends on hyperparameters like sampling variance and number of samples, which affect both stability and computational cost. Future work may explore adaptive tuning based on local obstacle density and real-world validation. Along with the Modified APF, the proposed algorithm for RF source seeking to compute the direction vector was also found to be quite accurate. However, this RF source seeking algorithm only works for the line of sight scenarios; if there is no line of sight, the drone will be confused about which angle reading to consider due to real-world factors like noise and multipath effects. Enhancing robustness using more antennas and estimation techniques, along with experiments in the real world, can be a focus for future development.

\appendix

\subsection{When the \textbf{\(det(A)=0\)} in RF Source Seeking Algorithm}

If the \( \det(A) \) becomes zero, the system \( x = A^{-1}b \) no longer yields a unique solution. This situation can result in two distinct cases: either there is no solution, or there are infinitely many solutions. 

\begin{figure}[htbp]
    \centering
    \includegraphics[width=0.95\linewidth]{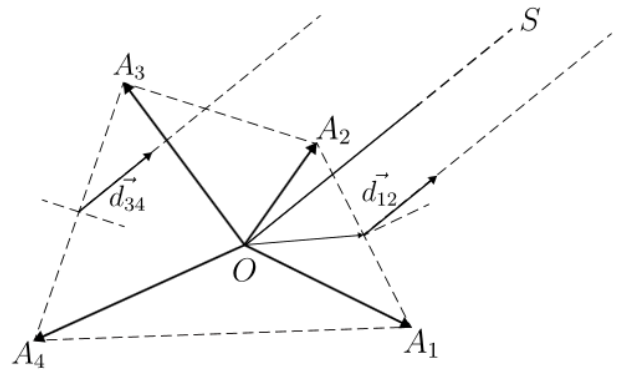}
    \caption{No Solution Case when \( \det(A) = 0 \).}
    \label{fig:5}
\end{figure}

In the case where no solution exists, the direction vectors corresponding to the AoA at \( A_1A_2 \) and \( A_3A_4 \) are parallel in the drone body coordinate system, as depicted in Figure \ref{fig:5}. This scenario indicates that the RF source is located at a significantly large distance relative to the drone, effectively appearing at infinity. Under such circumstances, the resultant direction vector, denoted as \( \vec{d}_s \), can be assumed to align with either of the calculated direction vectors, as they are equal in magnitude and direction:

\begin{align}
    \vec{d}_s = \vec{d}_{12} = \vec{d}_{34}.
\end{align}

\begin{figure}[htbp]
    \centering
    \includegraphics[width=0.95\linewidth]{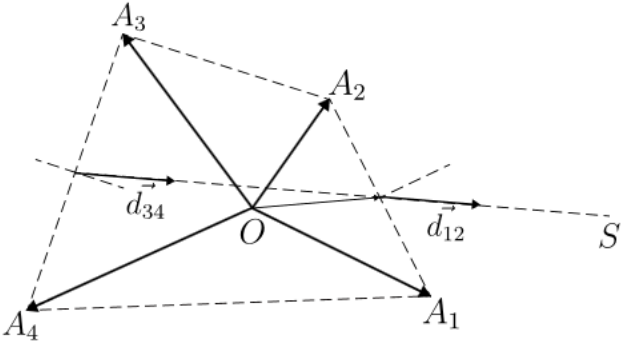}
    \caption{Infinite Solution Case when \( \det(A) = 0 \).}
    \label{fig:6}
\end{figure}

In the second case, where infinitely many solutions exist, the direction vectors remain parallel; however, the source, the midpoint of \( A_1 \) and \( A_2 \), and the midpoint of \( A_3 \) and \( A_4 \) are collinear. This configuration is illustrated in Figure \ref{fig:6}. Here, one direction vector points towards the other, as well as towards the RF source, within the drone body coordinate system. This implies that the scaled direction vectors intersect at infinitely many points, resulting in an infinite number of solutions to the system, in which case the resultant direction vector \( \vec{d}_s \) can again be assumed to align with either of the calculated direction vectors as shown above. 

Both cases arise due to the failure of the matrix \( A \) to provide a unique mapping, as indicated by its singularity (\(\det(A) = 0\)) where the resultant direction vector is given an approximate value such that in later iterations of running the algorithm the determinant will become non zero and the system will get a unique solution.

\subsection{Invalid Solution Set for RF Source Seeking Algorithm}

An additional scenario arises where the solution to the set of four equations is not valid, specifically when \( k_{12} \) or \( k_{34} \) becomes negative. This outcome corresponds to the reversal of the direction vector (\( \vec{d}_{12} \) or \( \vec{d}_{34} \)) such that it points in the opposite direction, which is not a physically valid solution. 

\begin{figure}[htbp]
    \centering
    \includegraphics[width=0.65\linewidth]{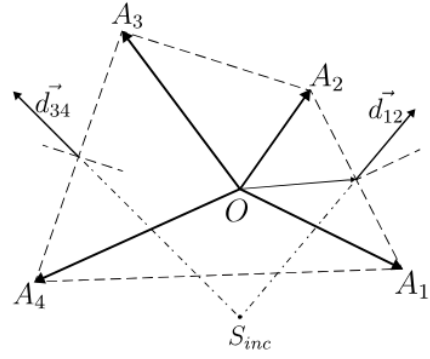}
    \caption{Diverging Direction vectors at \( A_{12} \) \& \( A_{34} \).}
    \label{fig:7}
\end{figure}

However, this situation does not pose a significant issue, as it occurs only when the direction vectors at both linear arrays are diverging, as illustrated in Figure \ref{fig:7}. Reversing these vectors mathematically would produce an intersection point to estimate an incorrect source position (\(S_{inc}\)). Nonetheless, such diverging vectors are not feasible in practical real-world scenarios because no single RF source can emit signals that would result in this configuration unless an extraordinary amount of noise is present, in which case techniques like MUSIC algorithm should be implemented before this to filter out the noise. 

Thus, the presence of negative scaling factors \( k_{12} \) or \( k_{34} \) can be disregarded as a practical concern.

\subsection{Three Antenna Configuration}

\begin{figure}[htbp]
    \centering
    \includegraphics[width=0.8\linewidth]{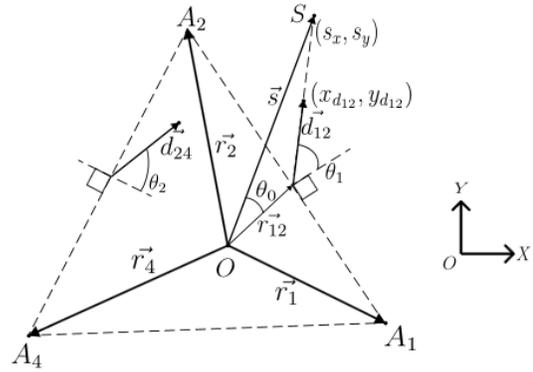}
    \caption{An Arbitrary Antenna Configuration with 3 Antennas.}
    \label{fig:8}
\end{figure}

The approach for deriving the source position can also be adapted to configurations with three antennas as discussed above. By making two antennas from different arrays coincide, the configuration effectively reduces to two distinct dipoles. For instance, by coinciding antennas \( A_2 \) and \( A_3 \), their position vectors become identical (\( \vec{r}_2 = \vec{r}_3 \)), allowing for substitution of all instances of \( \vec{r}_3 \) with \( \vec{r}_2 \) and corresponding values, as illustrated in Figure \ref{fig:8}.

Following the same derivation process as with four antennas, the resulting system of equations (\( Ax = b \)) is represented by:

\begin{align}
A = \begin{bmatrix}
1 & 0 & 0 & \frac{x_2 + x_4}{2} - x_{d_{24}} \\
0 & 1 & 0 & \frac{y_2 + y_4}{2} - y_{d_{24}} \\
1 & 0 & \frac{x_1 + x_2}{2} - x_{d_{12}} & 0 \\
0 & 1 & \frac{y_1 + y_2}{2} - y_{d_{12}} & 0
\end{bmatrix}
\& \quad
b = \begin{bmatrix}
\frac{x_2 + x_4}{2} \\
\frac{y_2 + y_4}{2} \\
\frac{x_1 + x_2}{2} \\
\frac{y_1 + y_2}{2}
\end{bmatrix}
\end{align}

The solution to this system is again \( x = A^{-1}b \), \& for a unique solution, determinant of \( A \) must satisfy \(\det(A) \neq 0\). This process ultimately yields both the coordinates and direction vector of the source.

In both the four-antenna and three-antenna configurations, the derivation employs exactly two linear arrays to determine the coordinates \((s_x, s_y)\). However, using more than two arrays is possible and transforms the system into an over-determined one. This setup, where more measurements are available than the minimum required, can be solved using estimation techniques like least squares estimation to obtain the coordinates in a way that minimizes error across all linear arrays.

\section*{DECLARATIONS}

\subsection*{Conflict of Interest}
The authors declare that there is no competing financial interest or personal relationship that could have appeared to influence the work reported in this paper.

\subsection*{Authors' Contributions}
Shahid and Aryan conducted the research and wrote the paper, Lakshmi reviewed the paper, and Anuj developed the idea, supervised the research, and reviewed the paper. 

\subsection*{Funding }
Not Applicable.

\bibliographystyle{unsrt}
\bibliography{references}


\relax 

\end{document}